\documentclass[conference]{IEEEtran}
\IEEEoverridecommandlockouts
\usepackage{cite}
\usepackage{amsmath,amssymb,amsfonts}
\usepackage{algorithm}
\usepackage{algpseudocode}
\usepackage{bbding}

\usepackage{graphicx}
\usepackage{textcomp}
\usepackage{xcolor}

\def\BibTeX{{\rm B\kern-.05em{\sc i\kern-.025em b}\kern-.08em
    T\kern-.1667em\lower.7ex\hbox{E}\kern-.125emX}}
\begin{document}

\title{PIONM: A Generalized Approach to Solving Density-Constrained Mean-Field Games Equilibrium under Modified Boundary Conditions\\

\thanks{This work was supported by the National Science and Technology Major Project (Grant No. 2022ZD0116401) and the Research Funding of Hangzhou International Innovation Institute of Beihang University (Grant No. 2024KQ161)\\
$^\dagger$ Corresponding Author: Wang Yao (yaowang@buaa.edu.cn) and Xiao Zhang (xiao.zh@buaa.edu.cn)}
}
\author{
\IEEEauthorblockN{
Jinwei Liu$^{1,3}$,
Wang Yao$^{2,3,4,5,\dagger}$,
Xiao Zhang$^{1,3,4,5,\dagger}$
}
\IEEEauthorblockA{
$^{1}$School of Mathematical Sciences, Beihang University, Beijing 100191, China
}
\IEEEauthorblockA{
$^{2}$School of Artificial Intelligence, Beihang University, Beijing 100191, China
}
\IEEEauthorblockA{
$^{3}$Key Laboratory of Mathematics, Informatics and Behavioral Semantics (LMIB), \\ Ministry of Education, Beijing 100191, China
}
\IEEEauthorblockA{
$^{4}$Hangzhou International Innovation Institute of Beihang University, Hangzhou 311115, China
}
\IEEEauthorblockA{
$^{5}$Zhongguancun Laboratory, Beijing 100094, China
}
}
\maketitle
\begin{abstract}
Neural network-based methods are effective for solving equilibria in Mean-Field Games (MFGs), particularly in high-dimensional settings. However, solving the coupled partial differential equations (PDEs) in MFGs limits their applicability since solving coupled PDEs is computationally expensive. Additionally, modifying boundary conditions, such as the initial state distribution or terminal value function, necessitates extensive retraining, reducing scalability. To address these challenges, we propose a generalized framework, PIONM (Physics-Informed Neural Operator NF-MKV Net), which leverages physics-informed neural operators to solve MFGs equations. PIONM utilizes neural operators to compute MFGs equilibria for arbitrary boundary conditions. The method encodes boundary conditions as input features and trains the model to align them with density evolution, modeled using discrete-time normalizing flows. Once trained, the algorithm efficiently computes the density distribution at any time step for modified boundary condition, ensuring efficient adaptation to different boundary conditions in MFGs equilibria. Unlike traditional MFGs methods constrained by fixed coefficients, PIONM efficiently computes equilibria under varying boundary conditions, including obstacles, diffusion coefficients, initial densities, and terminal functions. PIONM can adapt to modified conditions while preserving density distribution constraints, demonstrating superior scalability and generalization capabilities compared to existing methods.
\end{abstract}

\begin{IEEEkeywords}
Mean-Field Games, Physics-Informed Neural Operator, Normalizing Flow, Generalized Approach
\end{IEEEkeywords}

\section{Introduction}
Mean-Field Games (MFGs), independently introduced by Lasry and Lions in \cite{lasry2007mean} and Huang et al. in \cite{huang2006large}, offer a robust framework for tackling large-scale multi-agent problems. MFGs have been extensively applied in intelligent agent path planning, traffic flow optimization, and opinion dynamics. Low-dimensional MFGs problems are solvable through direct numerical methods. Nowadays, for high-dimensional problems involving coupled partial differential equations (PDEs), neural network-based algorithms have demonstrated high efficiency and have been recently applied to MFGs. For instance, \cite{lin2021alternating} reformulated MFGs as a Generative Adversarial Network (GAN) training problem, whereas \cite{ruthotto2020machine} proposed a Lagrangian approach to approximate agent states via sampling. Additionally, \cite{huang2023bridging} integrated MFGs constraints into the Schrödinger Bridge (SB) formulation of Optimal Transport (OT) to approximate MFGs equilibria. Most existing methods are trained under fixed initial and terminal boundary conditions, limiting their applicability to specific coefficient equations. Changes in boundary conditions often necessitate retraining, which is time-consuming and hampers the scalability of MFGs equilibrium solutions, restricting their practical applications.

In recent years, numerous efforts have been made to solve generalized MFGs equilibria. Among existing methods, \cite{perrin2022generalization} generalizes MFGs to different initial conditions using a grid and master policy. However, grid-based methods suffer from limited precision, restricting broader applications. The scalable learning approach in \cite{liu2024scalable} generalizes MFGs equilibria for traffic flow problems but relies on fixed and simplistic terminal value functions as well as specific forms of initial distributions, limiting its applicability to complex environment. Another approach, proposed in \cite{huang2024unsupervised}, generalizes MFGs equilibria using initial and terminal points; however, trajectory-based methods fail to compute intermediate density, which becomes ineffective in density-related loss functions. Furthermore, treating point sets as research objects overlooks key aspects of MFGs studies, fails to enforce density constraints, and significantly increases computational costs for large-scale equilibrium problems. These limitations restrict the broader application of MFGs in real-world scenarios.

Recently, neural operator-based networks have been developed to train models capable of predicting PDE outputs under various conditions in \cite{li2021fourier} and \cite{lu2021learning}. Notably, \cite{li2024physics} incorporates physical information into neural operator training via loss functions derived from mathematical equations, reducing training data requirements by leveraging PDE formulations. However, varying boundary conditions still impact the mathematical structure of the equations. Therefore, incorporating Physics-Informed Neural Operators (PINO) into existing generative models and coupling their training allows for enhanced generalization. This approach leverages mathematical equations to augment data-driven methods, enabling the solution of generalized MFGs equilibria.

In summary, we propose the Physics-Informed Neural Operator NF-MKV Net (PIONM), which integrates physics-informed neural operators with density-preserving Normalizing Flow (NF) generative models to solve MFGs equilibria under arbitrary boundary conditions. The proposed method encodes boundary conditions as inputs and trains the model to align them with density evolution flows modeled by discrete-time NF. Once trained, the algorithm can efficiently compute the density distribution at any time step of the system’s evolution for various boundary conditions, thereby enabling scalable solutions for MFGs equilibria. Compared to existing methods for generalized MFGs (Table \ref{tab3}), our approach enables multiparameter generalization of MFGs in continuous space based on density.

\begin{table}[htbp]
\caption{Capability Comparison with Existing Algorithms}
\begin{center}
\begin{tabular}{c c c c c }
\hline
\textbf{Desired}& Continuous&Generalizat-&Density&Multi-\\

\textbf{Features} &state space & ion capability &-based&parameter  \\
\hline
Ours& $\checkmark$& $\checkmark$& $\checkmark$ & $\checkmark$\\
Lin \cite{lin2021alternating}& $\checkmark$&$\times$& $\times$& $\times$\\
Perrin \cite{perrin2022generalization}& $\times$& $\checkmark$& $\checkmark$& $\times$\\
Liu \cite{liu2024scalable}& $\checkmark$& $\checkmark$& $\checkmark$& $\times$\\ 
Huang\cite{huang2024unsupervised}& $\checkmark$& $\checkmark$& $\times$& $\checkmark$\\ \hline
\end{tabular}
\label{tab3}
\end{center}
\end{table}

\textbf{Contribution} The primary contributions of this work are as follows:
\begin{itemize}
    \item We introduce PIONM, a method that leverages neural operators to solve MFGs equilibria under arbitrary boundary conditions. Our approach eliminates the need for retraining when boundary conditions change, significantly improving scalability.
    \item The proposed method incorporates NF as the backbone framework for solving fixed-coefficient MFGs. This ensures the preservation of density constraints during the solution process, aligning with the fundamental property of MFGs that the total density remains invariant.
    \item By combining physics-informed neural operators (PINO) with the NF generative model, the method encodes the initial and terminal conditions of MFGs as inputs. The discrepancy between the density flow derived from NF and the output of PINO is used as a loss to train PINO. During training, PINO provides reference approximations to NF, thereby accelerating the solution of MFGs equilibria with fixed coefficients.
    \item The method is applicable to scenarios involving changes in obstacle locations, diffusion coefficients of MFGs, and simultaneous variations in initial density distributions and terminal value functions.
\end{itemize}

\section{Connections among Generalized MFGs, NF-MKV Net and PINO}
\subsection{MFGs \& NF-MKV Net}
We now formalize the MFGs problem without considering common noise. For this purpose, we start with a complete filtered probability space $(\Omega,\mathcal{F},\mathbb{F}=(\mathcal{F}_t)_{0\leq t\leq T},\mathbb{P}))$ the filtration $\mathbb{F}$ supporting a $d-$dimensional Wiener process $\mathbf{W}=(W_t)_{0\leq t\leq T}$ with respect to $\mathbb{F}$ and an initial condition $\xi\in L^2(\Omega,\mathcal{F}_0,\mathbb{P};\mathbb{R}^d)$. This MFGs problem can be described as:

(i) For each fixed deterministic flow $\boldsymbol{\mu} = (\mu_t)_{0\leq t\leq T}$ of probability measures on $\mathbb{R}^d$, solve the standard stochastic control problem:

\begin{equation}
\begin{aligned}
    &\inf_{\alpha\in \mathbb{A}}J^\mu (\alpha) \quad \text{with}\quad J^\mu (\alpha)=\\ &\mathbb{E}[\int_0^T f(t,X^\alpha _t,\mu_t,\alpha_t)dt+g(X_T^\alpha,\mu_T)],
\end{aligned}
\label{con}
\end{equation}

subject to

\begin{equation}
    \left\{
    \begin{array}{l}
         dX_t^\alpha =b(t,X_T^\alpha,\mu_t,\alpha_t)dt+\sigma(t,X_T^\alpha,\mu_t,\alpha_t)dW_t, \\
         X_0^\alpha =\xi \qquad \qquad \qquad \qquad \qquad \qquad \quad t\in[0,T],
    \end{array}
    \right.
\end{equation}

(ii) Find a flow $\boldsymbol{\mu}=(\mu_t)_{0\leq t\leq T}$such that $\mathcal{L}(\hat{X}_T^{\boldsymbol{\mu}})=\mu_t$ for all $t\in[0,T]$, if $\hat{X}^{\boldsymbol{\mu}}$ is a solution of the above optimal control problem.

We can see that the first step provides the best response of a given player interacting with the statistical distribution of the states of the other players if this statistical distribution is assumed to be given by $\mu_t$. In contrast, the second step solves a specific fixed point problem in the spirit of the search for fixed points of the best response function.
\begin{figure*}[htbp]
\centerline{\includegraphics[width=\textwidth]{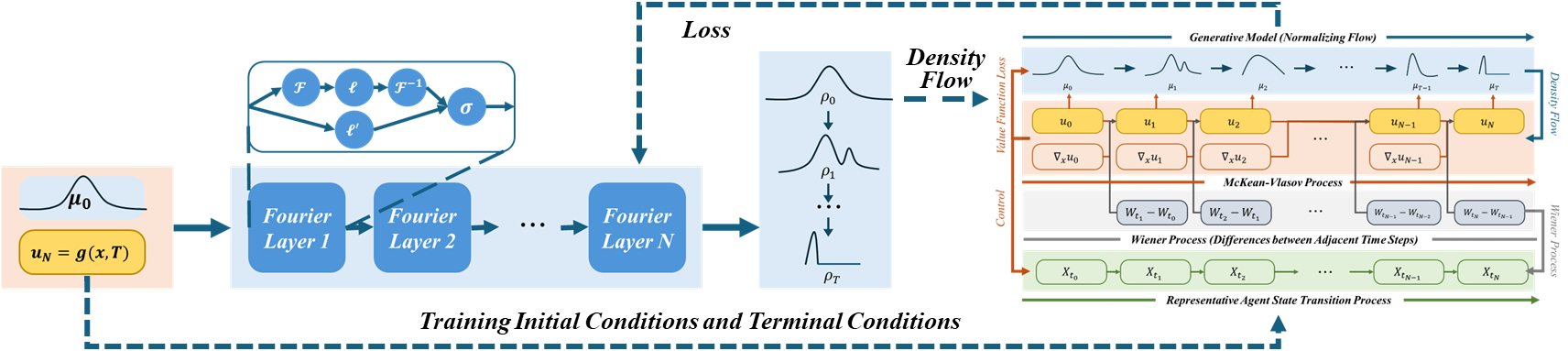}}
\caption{Construction of algorithm}
\label{cons}
\end{figure*}
Usually, the solution of MFGs is transformed into a set of coupled partial differential equations, namely the Hamilton-Jacobi-Bellman (HJB) and Fokker-Planck-Kolmogorov (FPK) equations equations, which respectively describe the evolution of the value function of the representative element and the density evolution of the group, as shown below:

\begin{equation}
    \begin{aligned}
& -\partial_t u-\sigma \Delta u+H(x, \nabla u)=f(x, \mu) \quad &&\text{(HJB)} \\
& \partial_t \mu-\sigma \Delta \mu-\operatorname{div}(\mu \nabla_p H(x, \nabla u))=0 \quad  &&\text{(FPK)}\\
& \mu(x, 0)=\mu_0, \quad u(x, T)=g(x, \mu(\cdot, T))
\end{aligned}\label{mfg}
\end{equation}

in which, $u:\mathbb{R}^n\times [0,T]\rightarrow \mathbb{R}$ is the value function to guide the agents make decisions; $H:\mathbb{R}^n\times \mathbb{R}^n\rightarrow\mathbb{R}$ is the Hamiltonian, which describes the physics energy of the system; $\mu(\cdot,t)\in \mathcal{L}(\mathbb{R}^n)$is the distribution of agents at time t, $f:\mathbb{R}^n\times \mathcal{L}(\mathbb{R}^n)\rightarrow \mathbb{R}^n$ denotes the loss during process; and $g:\mathbb{R}^n\times \mathcal{L}(\mathbb{R}^n)\rightarrow \mathbb{R}^n$ is the terminal condition, guiding the agents to the final distribution, while $\sigma$ represents the diffusion coefficient, controlling the stochasticity of agent dynamics.

Let assumption \textbf{MFGs Solvability HJB} be in force in [5]. Then, for any initial condition $\xi\in L^2(\Omega,\mathcal{F}_0,\mathbb{P};\mathbb{R}^d)$, the \textit{McKean-Vlasov FBSDEs}:

\begin{equation}
    \left\{\begin{array}{l}
d X_t= b(t, X_t, \mathcal{L}(X_t), \hat{\alpha}(t, X_t, \mathcal{L}(X_t), \sigma) dt +\sigma d W_t \\
d Y_t= -f(t, X_t, \mathcal{L}(X_t), \hat{\alpha}(t, X_t, \mathcal{L}(X_t), \sigma) d t +Z_t \cdot d W_t
\end{array}\right.
\label{fbsde}
\end{equation}
for $t\in [0,T]$, with $Y_T=g(X_T,\mathcal{L}(X_T))$ as terminal condition and $\sigma$ as $\sigma(t, X_t, \mathcal{L}(X_t))^{-1 \dagger} Z_t)$, is solvable. Moreover, the flow $(\mathcal{L}(X_T))_{0\leq t\leq T}$ given by the marginal distributions of the forward component of any solution is an equilibrium of the MFGs problem associated with the stochastic control problem \ref{con}.

We analyze the optimal control step in the previously formulated MFGs problems. Probabilists approach these optimal control problems using a two-step method. We assume the input $\boldsymbol{\mu}=(\mu_t)_{0\leq t\leq T}$ is deterministic and fixed, allowing us to determine the optimal reaction. Given this fixed decision, we then solve for the optimal probability measure flows. Alternately seeking for the optimal control, the MFGs equilibrium can be finally derived.

We use NF-MKV Net throughout the paper to refer to the constraint-preserving neural network framework for solving MFGs. In \cite{liu2025nfmkv}, the alternating training model NF-MKV Net was introduced to solve MFGs equilibria. This model combines the NF framework with a time-series neural network, linking states and strategies to solve the fixed-point problem of McKean-Vlasov type Forward-Backward Stochastic Differential Equations (MKV FBSDEs), which correspond to the equilibria of MFGs. MKV FBSDEs offer the advantage of encapsulating both optimization and interaction within a single coupled FBSDE, removing the need to separately reference the HJB and FPK equations.

The NF framework uses neural networks to generate flows, enforcing constraints on each density transition function to define the density distribution at specific time steps. The NF-derived density flow is coupled with the MFGs value function. The value function constrains the neural networks generating the density flow via the HJB equation, while its gradient updates depend on the current marginal density flow.

\subsection{Generalized MFGs \& PINO}
The MFGs problem is formulated as the evolution of an initial density, $\mu_0$, constrained by $\int_0^T f(x,t)dt + g(x,T)$, to minimize both process and terminal value funstion. Variations in boundary conditions lead to corresponding changes in the governing equations. The system's initial density is given by $\mu_0$, environmental dynamics by $f(x,t)$, and the terminal state by $g(x,T)$. Furthermore, individual behavior is influenced by the diffusion coefficient $\sigma$.

The generalized MFGs equilibrium solution adapts to boundary condition changes while minimizing complete re-computation, maintaining consistency with the equilibrium. Achieving this requires effective boundary condition encoding, enabling their finite representation.

In specific scenarios, such encoding simplifies the methodology. In traffic flow problems, the initial density can be represented by wave functions or normal distributions, while terminal conditions are set to zero or specific states. In Crowd Motion problems, the initial density can be modeled as normal or uniform distributions with specified mean and variance. Terminal conditions are defined by the Euclidean distance to a target point, while environmental obstacles are pre-processed using convex hull representations.

In Generalized MFGs, the objective is to efficiently compute equilibria for various boundary conditions, including arbitrary initial distributions and terminal value functions. PINO provides an effective approach for achieving this generalization. PINO, based on the Fourier Neural Operator (FNO), learns mappings between infinite-dimensional function spaces using the Fourier transform. This transformation projects inputs into a high-dimensional space, enhancing the modeling of variable relationships. By embedding control system laws into its loss function, PINO applies physics-informed regularization, ensuring predictions align with known dynamics.

Unlike traditional neural networks embedded with mathematical principles, such as NF-MKV Net, PINO encodes and generalizes input information through the Fourier transform within the FNO framework. This enables PINO to overcome challenges associated with fixed-coefficient methods, which struggle to propagate information effectively under varying boundary conditions. The transformation projects boundary conditions into a higher-dimensional space, endowing PINO with strong generalization capabilities for solving MFGs equilibria under diverse boundary conditions. By providing a scalable learning framework, PINO eliminates the need to retrain multiple equations for each new boundary condition, addressing a major limitation of traditional methods.
\section{Methodology: PIONM}
We propose PIONM, a neural operator-based method for solving MFGs equilibria with arbitrary boundary conditions. PIONM provides generalized solutions to MFGs equilibria across different initial conditions and terminal value functions without retraining. The solution is represented as the system’s density evolution over time steps. This approach utilizes the NF framework for fixed-coefficient MFGs, preserving density distribution constraints and ensuring mass conservation throughout the solution process.

PIONM integrates PINO with the generative NF model, using initial and terminal conditions of MFGs equilibria as inputs. The loss function is defined as the discrepancy between the density flow from NF and the PINO output, guiding effective training of PINO.

Initially, boundary conditions are encoded: the initial distribution is defined by the center and diffusion coefficients of a Gaussian distribution, terminal value functions are given as target coordinates, and obstacles are represented by their positions and radii. he encoded boundary conditions serve as inputs for training PIONM. During training, PINO generates approximate reference solutions for NF, accelerating the computation of MFGs equilibria with fixed coefficients. After training, PIONM efficiently solves MFGs equilibria for any encodable boundary conditions. Figure \ref{cons} presents the PIONM framework.

\subsection{Solving MFGs with NF-MKV Net}

We use NF-MKV Net to solve fixed-coefficient MFGs. 

\begin{figure}[htbp]
\centerline{\includegraphics[width=0.5\textwidth]{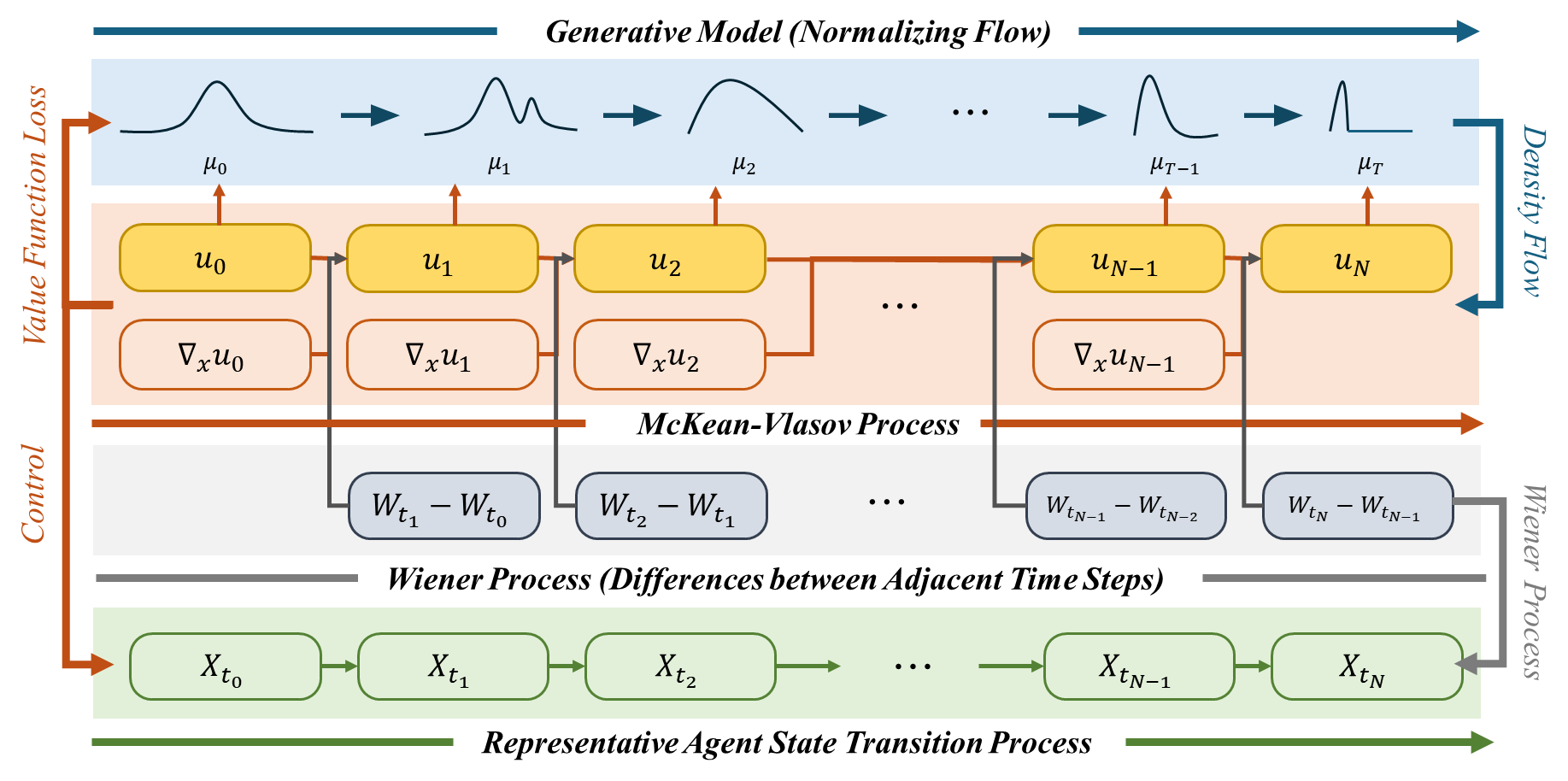}}
\caption{NF-MKV Net}
\label{fix}
\end{figure}

We reformulate the stochastic equations of the MFGs using the MKV FBSDE and approximate the gradient of the value function with a neural network, which effectively solves the curse of dimensionality problem in traditional numerical methods. Second, to address the challenge of distributional coupling, we use a neural network architecture to model the agent's state density distribution, alternately training the unknown transition process with the value function. Figure \ref{fix} illustrates the framework of NF-MKV Net.

After discretization of the process, all value functions are connected by summing over $t$. The network uses the generated density flows $\boldsymbol{\mu}$ and $W_{t_n}$ as inputs and produces the final output $\hat{u}$, approximating $u(x, T) = g(x, \mu(\cdot, T))$. This approximation defines the expected loss function by comparing the difference between the two functions for $\{x_i\}_{i=1}^N \sim z=\hat{\mu}_T$:
\begin{equation}
\begin{aligned}
    l_{\textit{MKV}}=&\sum\|g(z,\hat{\mu}_T)-\hat{u}(\theta,z))\|^2\\ =&-\frac{1}{N}\sum\nolimits _{i=1}^N \|g(x_i,\hat{\mu}_T)-\hat{u}(\theta,x_i))\|^2.
\end{aligned}
\label{nll1}
\end{equation}

At each time step $t = t_n$, the MFGs system satisfies the HJB equation. Thus, samples $\{x_i\}_{i=1}^M$ from $\mu_t$ at each time step can be used in the HJB equation to compute the loss,

\begin{equation}
\begin{aligned}
     l_{\text {HJB}}=\frac{1}{N}\frac{1}{M}\sum\nolimits_{n=1}^N \sum\nolimits_{i=1}^M \| \partial_t u\left(x_i, t_n\right)+\nu \Delta u\left(x_i, t_n\right) \\ -H\left(\nabla_x u\left(x_i, t_n\right)\right)+f\left(x_i, t_n\right) \|^2
\end{aligned}
\label{lhjb}
\end{equation}
where 
\begin{equation}
\begin{aligned}
    &\left(\{x_i\}_{i=1}^M,t_n\right)\sim \mu_{t_n}\approx \mu_{t_n}(\phi)\\ &=r_{t_{n}}(x;\phi_n)\circ r_{t_{n-1}}(x;\phi_{n-1})\circ \cdots \circ r_{t_1}(x;\phi_1) \circ \mu_0.
\end{aligned}
\end{equation}

Additionally, the NF method must match the terminal density condition, so the terminal value function $l_T$ is also included in the loss calculation. If the terminal condition $g$ is explicitly defined, the corresponding optimal density $\hat{\mu}_T (\phi) = \mathbf{f}(x; \Phi) \circ \mu_0$ can serve as the target distribution for NF. $\hat{\mu}_T(\phi)$ generated by NF is used to compute the terminal value function $l_T$. For $\{x_i\}_{i=1}^N \sim \hat{\mu}_T(\Phi)$:

\begin{equation}
    l_{\textit{T}}=\frac{1}{N}\sum\nolimits _{i=1}^N g(x_i).\label{nll2}
\end{equation}

\subsection{Sovling generalized MFGs using PINO}
We introduce the PINO framework to capture population dynamics and the system value function for solving equilibria in generalized MFGs. The equilibria of MFGs are influenced by initial densities, terminal value functions, and environmental information. Fig. \ref{cons} illustrates the workflow of the proposed framework. The PINO module employs the Fourier Neural Operator (FNO) to model the temporal dynamics of density distribution $\{\mu_i\}_{i=1}^N$. Given specific boundary conditions, the PINO framework computes the time- and state-dependent individual density distributions, which are then passed to the NF-MKV Net module. This module uses the boundary conditions and the transmitted density distributions for initialization to solve the fixed-coefficient MFGs equilibria. The difference between the converged solution and the density distributions computed by PINO is used as a loss to optimize the PINO network. This approach enables generalized solutions to MFGs equilibria without requiring pre-specified data inputs.

\begin{figure}[htbp]
\centerline{\includegraphics[width=0.4\textwidth]{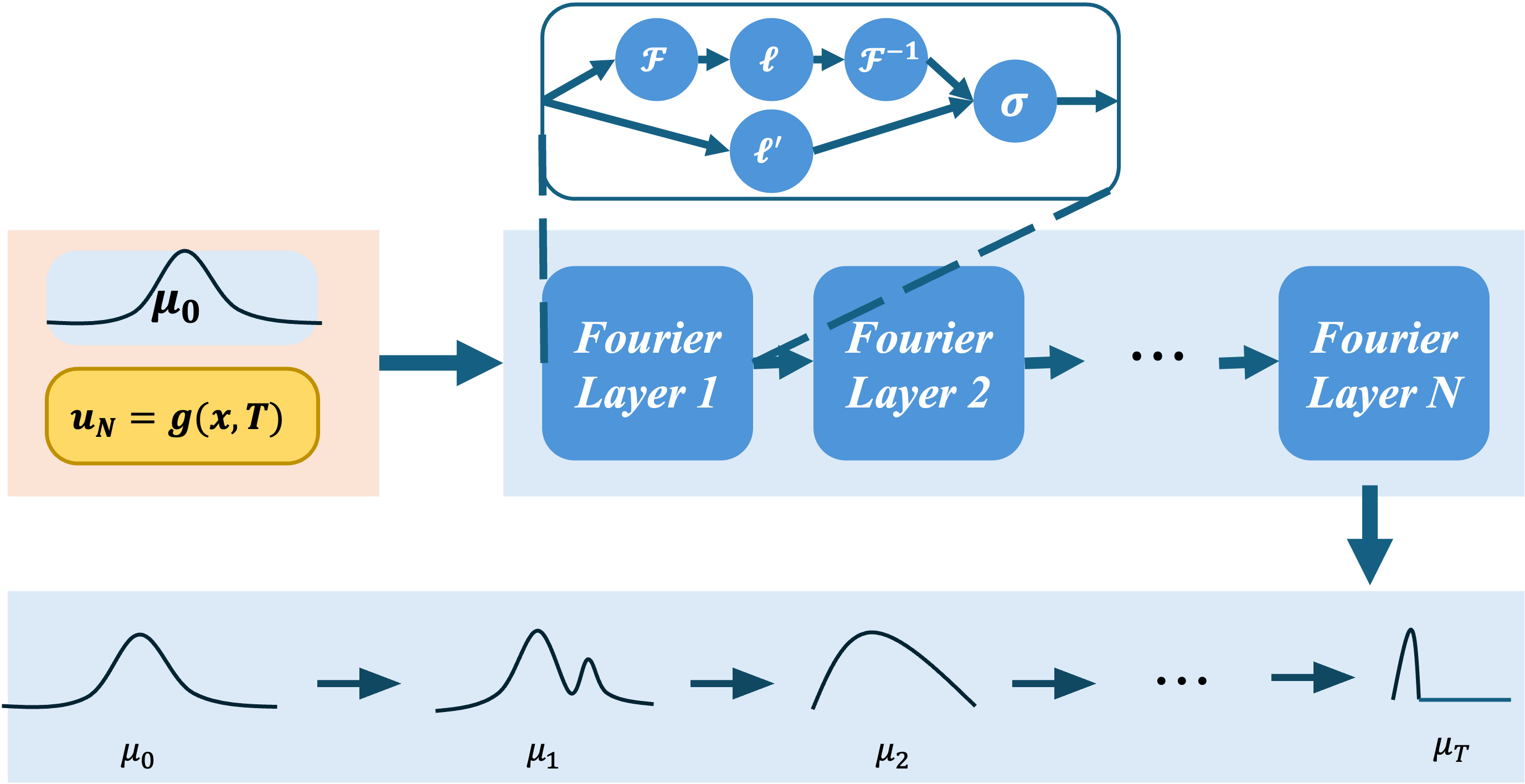}}
\caption{PINO solving}
\label{pino}
\end{figure}

Assuming the solution operator is denoted as $\mathcal{G}_\theta$  and the boundary conditions as $\mathcal{L}{\text{con}}$, we define the operator space as $\mathcal{G}_\theta : (\mathcal{L}{\text{con}}, x) \to \mathbb{P}^N(x)$, which maps boundary conditions to the individual density distribution values for all discrete time points in a given state. The training process minimizes the loss function:
\begin{equation}
\begin{aligned}
     l_{\text{PINO}}=\frac{1}{N}\frac{1}{M}\sum\nolimits_{n=1}^N \sum\nolimits_{i=1}^M \| \mathcal{G}_\theta(\mathcal{L}{\text{con}},x_i,t_n)-\mu_{t_n}(x_i)\|^2
\end{aligned}
\label{lpino}
\end{equation}
 
where minimizing $l_{\text{PINO}}$  yields the solution operator $\mathcal{G}_\theta$. For any input consistent with the encoded boundary conditions and a specific state, $\mathcal{G}_\theta$ computes the density distribution values of individuals at discrete time points. This enables the generalized solution of MFGs equilibria density evolution flows, corresponding to the equilibria of MFGs. Algorithm \ref{alg1} presents the pseudo-code of the model.

\begin{algorithm}
    \caption{PIONM}
    \label{alg1}
    \begin{algorithmic}
        \Require{$\sigma$ diffusion parameter, $H$ Hamiltonian}
        \Ensure{$\mathcal{G}_\theta$}
        \While{not converged}
        \State Random boundary conditions $\mathcal{L}{\text{con}}$
        \State Initialize $\{\mu_{t_n}\}_{i=1}^N=\mathcal{G}_\theta(\mathcal{L}{\text{con}},x_i,t_n)$
        \State Initialize MFGs$(\mu_0,g_T,f)\gets \mathcal{L}{\text{con}}$
        \While{Not Solving MFGs$(\mu_0,g_T,f)$}
        \State \textbf{NF-MKV Net:}
        \State \textbf{Train} $u(0,x|\theta_0)$ and $[\partial_x u(x,t) | \theta_n]^T\sigma$
        \State Back-propagate the loss $l_{\textit{MKV}}$ to $\theta$ weights.
        \State \textbf{Train} $r_n(\phi_n)$
        \State Back-propagate the loss $l_\text{NF}=l_{\text {HJB}}+l_{\textit{T}}$ to $\phi$ weights.
        \EndWhile
        \State Sample batch $\left(\{x_i\}_{i=1}^M,t_n\right)\sim \mu_{t_n}$
        \State Calculate $l_{\text{PINO}}$ by (\ref{lpino})
        \State Back-propagate the loss $l_{\text{PINO}}$ to $\mathcal{G}_\theta$ weights.
        \EndWhile
    \end{algorithmic}
\end{algorithm}

\section{Numerical Experiment}
We apply PIONM to generalized MFGs instances and present the numerical results in two parts. The first part demonstrates PIONM as an effective method for solving generalized MFGs equilibrium involving density distributions. The second part highlights the accuracy of PIONM Net in comparison to other algorithms.

\subsection{Solving generalized MFGs with PIONM}

In this example, a dynamically formulated MFGs problem, the Crowd Motion problem, is constructed in dimensions $d=2$ to demonstrate the applicability of PIONM. We set the problems as in (\ref{mfg}) with forms:

\begin{equation}
\begin{aligned}
    &f(x,\mu) = f_{\text{in}} + \int _{\mathbb{R}^n}\left( \mathbf{e}^{-|x - x_{\text{o}}|^2} + \mathbf{e}^{-||x - x_{\text{o}}|^2 - s_{\text{safe}}|^2}\right) d\mu(x).\\
    &\mu(x,0)= \mathcal{N}(x_0, \sigma^2_0), \qquad g(x,T)=\int _{\mathbb{R}^2} {|x - x_T|^2} d\mu(x).
\end{aligned}
\label{cm2d}
\end{equation}

\subsubsection{Obstacle Change Scenarios}
In the Crowd Motion scenario, we fixed the initial density distribution as $\mu(x,0)= \mathcal{N}((-7,0), 0.2^2)$ and the terminal value function as $g(x,T)=\int _{\mathbb{R}^2} {|x - (7,0)|^2} d\mu(x)$. Obstacles, represented by the encoding $(x_o, y_o, R)$, were modeled as circular shapes with centers at $(x_o, y_o)$ and radius $R$ to serve as boundary conditions. This encoding generalized the boundary conditions.

\begin{figure}[htbp]
\centerline{\includegraphics[width=0.5\textwidth]{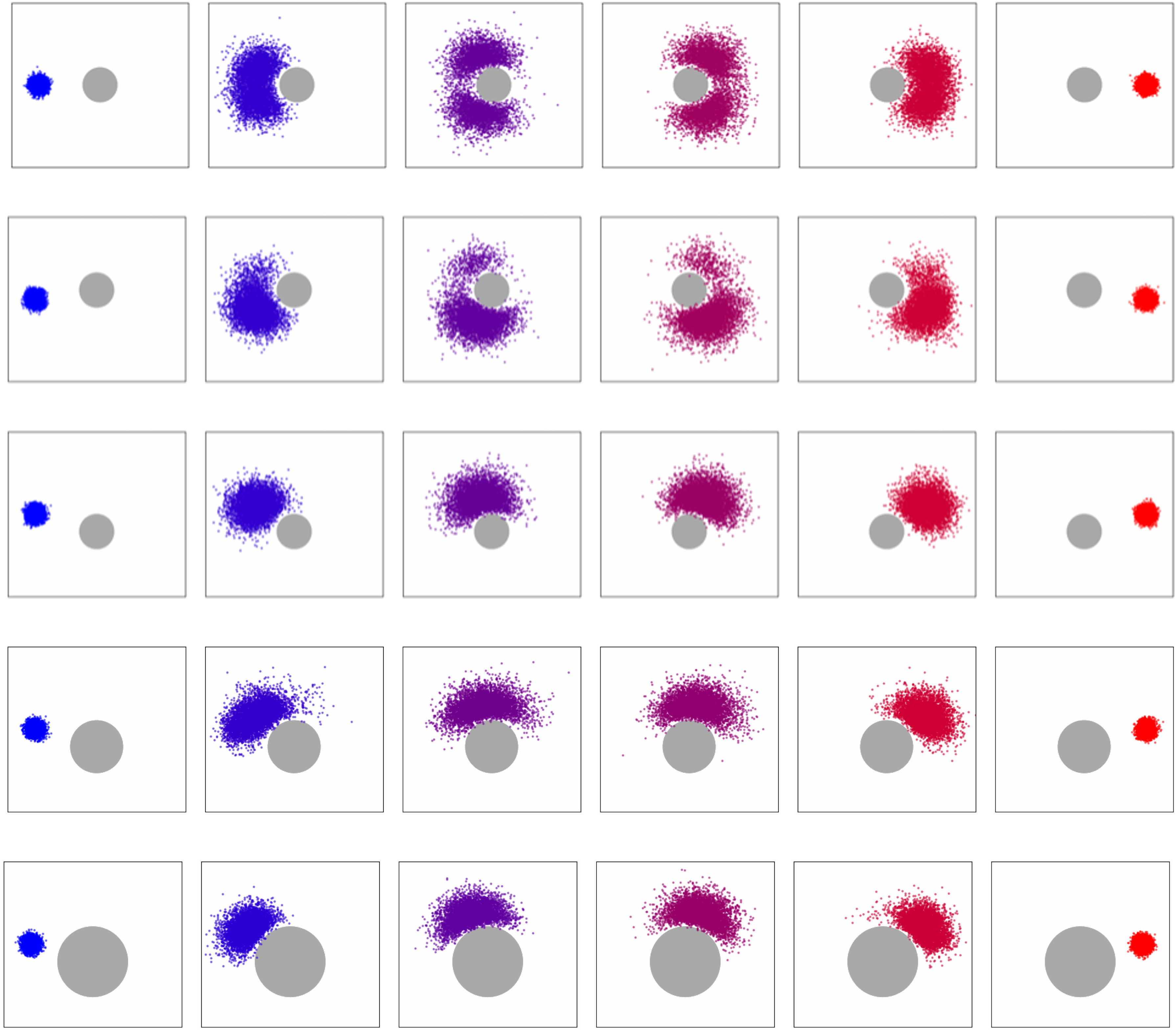}}
\caption{Obstacle Change}
\label{res-1}
\end{figure}

The obstacles were set at $(0,0,2)$, $(0,1,2)$, $(0,-2,2)$, $(0,-2,3)$, $(0,-2,4)$, and the training results are shown in Fig. \ref{res-1}. Each column illustrates the trajectories of 1000 agents in two dimensions under different obstacle configurations at time steps $0, 0.2T, 0.4T, 0.6T, 0.8T, T$. Simulation results show that PIONM effectively guides the initial Gaussian density to avoid obstacles with varying boundary conditions while evolving toward the terminal state that minimizes the loss. The framework ensures obstacle avoidance and collision prevention within the population.

\subsubsection{Diffusion Parameters Change Scenarios}
In this scenario, we maintained the initial density as $\mu(x,0)= \mathcal{N}((-10,0), 1^2)$, the terminal value function as $g(x,T)=\int _{\mathbb{R}^2} {|x - (10,0)|^2} d\mu(x)$, and introduced two fixed elliptical obstacles forming a narrow passage. The diffusion coefficient $\sigma$ in the stochastic Wiener process was varied to generalize boundary conditions.

\begin{figure}[htbp]
\centerline{\includegraphics[width=0.5\textwidth]{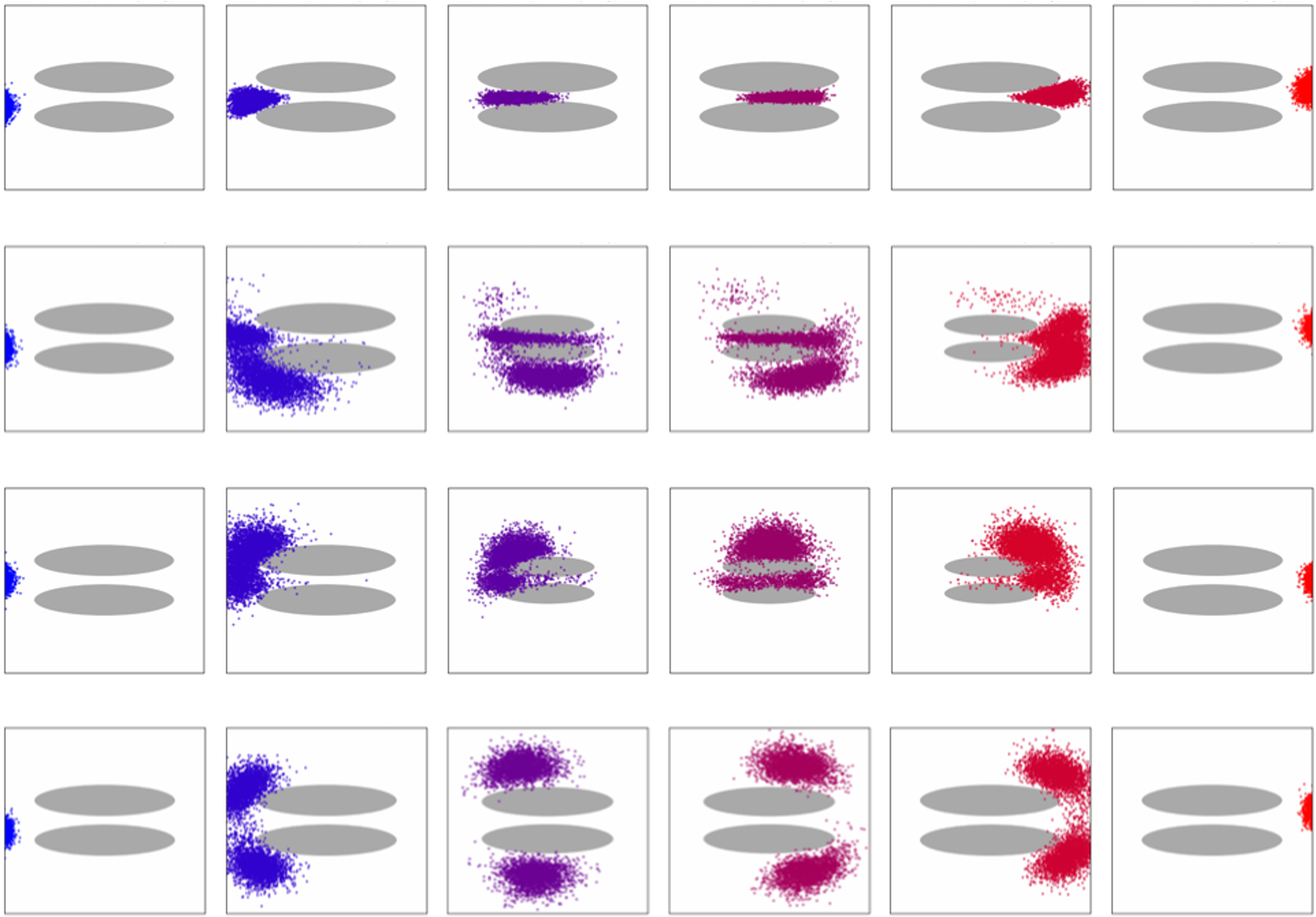}}
\caption{Diffusion Parameters Change}
\label{res-2}
\end{figure}

The diffusion coefficients were set to $0.2$, $1$, and $2$, and the training results are shown in Fig. \ref{res-2}. Each column shows the trajectories of 1000 agents under different diffusion coefficients at time steps $0, 0.2T, 0.4T, 0.6T, 0.8T$, and $T$. Specifically, the diffusion coefficients were set to $0.2$ (Column 1), $1$ (Columns 2 and 3), and $2$ (Column 4). The simulation results show that PIONM guides the population from the initial density to the terminal state while minimizing loss, completing the path-planning task, avoiding obstacles, and preventing inter-population collisions.

Notably, PIONM exhibits different behaviors under varying diffusion coefficients in the same environment. With smaller coefficients, agents tend to move through the narrow passage between the two obstacles. In contrast, with larger coefficients, agents perceive the obstacles as a single entity and bypass them entirely. Additionally, different training runs of PIONM exhibit varying navigation strategies under the same diffusion coefficients (e.g., Columns 2 and 3). Due to the inherent symmetry of the framework, both behaviors constitute valid solutions to the MFGs equilibrium.

\subsubsection{Initial and Terminal Conditions Change Scenarios}
In this case, we fixed the diffusion coefficients and obstacle configurations while varying the initial density distribution $\mu(x,0)= \mathcal{N}((x_0,y_0), \sigma_0^2)$ and terminal value functions $g(x,T)=\int _{\mathbb{R}^2} {|x - (x_T,y_T)|^2} d\mu(x)$. The initial density was encoded as a Gaussian distribution with mean $(x_0,y_0)$ and variance $\sigma_0$, and the terminal value function, relevant for path-planning tasks, was encoded as the destination coordinates $(x_T,y_T)$. These were used to generalize boundary conditions.

\begin{figure}[htbp]
\centerline{\includegraphics[width=0.5\textwidth]{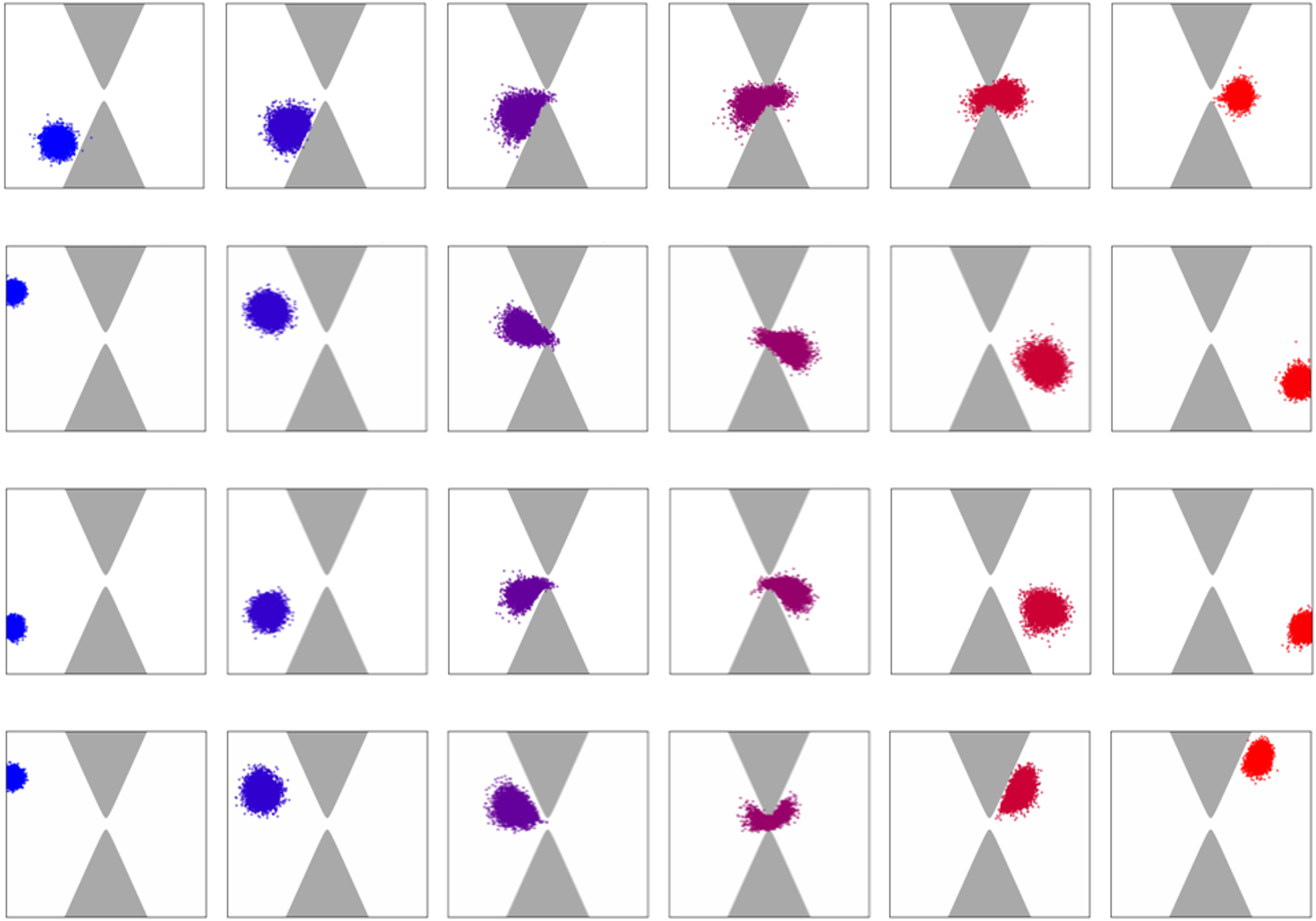}}
\caption{Initial and Terminal Conditions Change}
\label{res-3}
\end{figure}

The training results are presented in Fig. \ref{res-3}. Each column represents the trajectories of 1000 agents under different initial boundary conditions at time steps $0, 0.2T, 0.4T, 0.6T, 0.8T, T$. The initial densities and terminal value functions were set as in the table. The results demonstrate that PIONM effectively evolves the population from the initial density towards the terminal state minimizing loss, fulfilling the path-planning objectives while ensuring safe navigation and avoiding inter-agent as well as agent-obstacle collisions.

\begin{table}[htbp]
\caption{Initial and Terminal Conditions}
\begin{center}
\begin{tabular}{c c c c}
\hline
\textbf{Figure}&\multicolumn{3}{c}{\textbf{Initial and Terminal Conditions}} \\
\cline{2-4} 
\textbf{Column} & $(x_0,y_0)$& $\sigma_0$& $(x_T,y_T)$ \\
\hline
Col 1& $(-5,-5)$& $1$& $(3,0)$ \\
Col 2& $(-10,5)$& $0.2$& $(10,-5)$ \\
Col 3& $(-10,-5)$& $0.2$& $(10,-5)$ \\
Col 4& $(-10,5)$& $0.2$& $(5,5)$ \\ \hline
\end{tabular}
\label{tab1}
\end{center}
\end{table}

\subsection{Compared with other Methods}

To verify the generalization of algorithm, we compare PIONM with existing MFGs solving methods, including the distribution-based \textbf{RL-PIDL} method proposed by \cite{chen2023hybrid} and the high-dimensional neural network-based \textbf{APAC-Net} by \cite{lin2021alternating}, and \textbf{Scalable} Learning Method by \cite{liu2024scalable}.

\textbf{Solving Time.} We compared our approach with existing single-instance neural MFGs solvers to evaluate its accuracy and efficiency. All methods were tested under the same initial distribution and terminal value function settings.

\textbf{Collision Avoidance Success Rate.} In the Crowd Motion scenario, the collision avoidance rate is a critical metric for assessing the safety of an algorithm. Collision avoidance encompasses two aspects: avoidance between individuals and avoidance between individuals and obstacles. We calculated the overall collision avoidance rate by combining these two components to evaluate the safety performance of different algorithms.

\textbf{Invariance of Distribution Volume.} Additionally, since our method is based on the NF-MKV Net framework, it inherently preserves density during the evolution of the population system. To verify this property, we performed approximate integration over dynamic regions, a technique widely used in density estimation (e.g., \cite{WOS:000378444200012}). By generating grids in specified regions and numerically integrating the specified probability distribution within those regions, the results should closely approach 1, reflecting the invariance of the distribution volume.

Table \ref{tab2} records the solution time, collision avoidance rate, and distribution volume invariance for each method across all test scenarios. Analysis shows that the PIONM learning method improves inference speed by several orders of magnitude without sacrificing solution quality. Compared to existing methods based on PINO, our approach demonstrates better performance in terms of solution time.

\begin{table}[htbp]
\caption{Compared with other Methods}
\begin{center}
\begin{tabular}{c c c c c}
\hline
\textbf{Comparison}&\multicolumn{3}{c}{\textbf{Algorithms}} \\
\cline{2-5} 
\textbf{Aspects} & PIONM & RL-PIDL & APAC-NET & Scalable\\
\hline
Solving Time& 3s& 1576s & 10357s& 58s \\
Success Rate& $98.75\%$ & $/^*$& $96.63\%$ & $/^*$ \\
Volume Diff.$^{**}$& $-1.25$& $-0.34$& $/^{***}$& $0.27$\\\hline
\multicolumn{5}{l}{$^{\mathrm{*}}$Traffic Flow Methods do not involve collision.}\\
\multicolumn{5}{l}{$^{\mathrm{**}}$$\log $ of $\mu$ integral difference from $1$. By generating grids in specified}\\
\multicolumn{5}{l}{ regions and numerically integrating the specified probability distribution}\\
\multicolumn{5}{l}{ within those regions, the results should closely approach 1.}\\
\multicolumn{5}{l}{$^{\mathrm{***}}$Sample-based Method does not involve distribution.}\\
\end{tabular}
\label{tab2}
\end{center}
\end{table}

The proposed algorithm ensures a high collision avoidance success rate while achieving rapid solutions, which is of significant importance for safety in practical applications. Furthermore, the invariance of distribution volume enables the method to handle density-related MFGs equilibria with higher precision.

\subsection{Error Analysis}
The error due to discretized MKV FBSDEs is negatively correlated with the number of temporal discretizations $N$, i.e., $O(\delta_u)\sim O(\frac{1}{N})$. Therefore, the denser the temporal discretization splits, the smaller the resulting error in discretizations. Meanwhile, The training loss can be expressed as the solution loss of the discretized MFGs, and the error is caused by the parameterized Neural Network. Below shows the detail of error analysis.

Meanwhile, PINO $\mathcal{G}_\theta$  is a neural network parameterized by $\theta$, with its functional complexity measured by the Rademacher complexity or covering number of the hypothesis class. The generalization error can be estimated as
\begin{equation}
    \delta_{\text{PINO}} \leq C \cdot \mathcal{R}_N(\mathcal{H}) + O\left(\frac{1}{\sqrt{N}}\right),
\end{equation}
where $N$  is the number of training samples. As $N$  increases, the sampling error term $O(1/\sqrt{N})$  decreases, indicating that more training data significantly reduces the generalization error. This implies that the generalization error of PINO is strongly dependent on both the quantity and distribution of training data. The relationship $\delta_{\text{PINO}} \sim O(1/\sqrt{N})$  demonstrates that increasing the sample size reduces the error.

\section*{Conclusion}
This paper presents a generalized framework, called PIONM, for solving MFGs equations using PINO. PIONM employs neural operators to compute MFGs equilibria under arbitrary boundary conditions. The method is applicable to scenarios involving changes in obstacle locations, diffusion coefficients of MFGs, and simultaneous variations in initial density distributions and terminal value functions. The method represents boundary conditions as input features and trains the model to align them with density evolution. By integrating PINO with the NF generative model, the method incorporates the initial and terminal conditions of MFGs as inputs. The difference between the density flow obtained from NF and the output of PINO serves as the loss function for training. During training, PINO offers reference approximations to NF, accelerating the computation of MFGs equilibria with fixed coefficients. The proposed method is validated in diverse scenarios, demonstrating its effectiveness and applicability to MFGs problems with varying conditions.

\bibliographystyle{IEEEtran}
\bibliography{reff}

\begin{thebibliography}{10}
\providecommand{\url}[1]{#1}
\csname url@samestyle\endcsname
\providecommand{\newblock}{\relax}
\providecommand{\bibinfo}[2]{#2}
\providecommand{\BIBentrySTDinterwordspacing}{\spaceskip=0pt\relax}
\providecommand{\BIBentryALTinterwordstretchfactor}{4}
\providecommand{\BIBentryALTinterwordspacing}{\spaceskip=\fontdimen2\font plus
\BIBentryALTinterwordstretchfactor\fontdimen3\font minus
  \fontdimen4\font\relax}
\providecommand{\BIBforeignlanguage}[2]{{%
\expandafter\ifx\csname l@#1\endcsname\relax
\typeout{** WARNING: IEEEtran.bst: No hyphenation pattern has been}%
\typeout{** loaded for the language `#1'. Using the pattern for}%
\typeout{** the default language instead.}%
\else
\language=\csname l@#1\endcsname
\fi
#2}}
\providecommand{\BIBdecl}{\relax}
\BIBdecl

\bibitem{lasry2007mean}
J.-M. Lasry and P.-L. Lions, ``{Mean Field Games},'' \emph{Japanese journal of
  mathematics}, vol.~2, no.~1, pp. 229--260, 2007.

\bibitem{huang2006large}
M.~Huang, R.~P. Malham{\'e}, and P.~E. Caines, ``{Large Population Stochastic
  Dynamic Games: Closed-Loop McKean-Vlasov Systems and the Nash Certainty
  Equivalence Principle},'' \emph{JOURNAL OF SYSTEMS SCIENCE \& COMPLEXITY},
  2006.

\bibitem{lin2021alternating}
A.~T. Lin, S.~W. Fung, W.~Li, L.~Nurbekyan, and S.~J. Osher, ``{Alternating the
  Population and Control Neural Networks to Solve High-Dimensional Stochastic
  Mean-Field Games},'' \emph{Proceedings of the National Academy of Sciences},
  vol. 118, no.~31, p. e2024713118, 2021.

\bibitem{ruthotto2020machine}
L.~Ruthotto, S.~J. Osher, W.~Li, L.~Nurbekyan, and S.~W. Fung, ``{A Machine
  Learning Framework for Solving High-Dimensional Mean Field Game and Mean
  Field Control Problems},'' \emph{Proceedings of the National Academy of
  Sciences}, vol. 117, no.~17, pp. 9183--9193, 2020.

\bibitem{huang2023bridging}
H.~Huang, J.~Yu, J.~Chen, and R.~Lai, ``Bridging mean-field games and
  normalizing flows with trajectory regularization,'' \emph{Journal of
  Computational Physics}, vol. 487, p. 112155, 2023.

\bibitem{perrin2022generalization}
S.~Perrin, M.~Lauri{\`e}re, J.~P{\'e}rolat, R.~{\'E}lie, M.~Geist, and
  O.~Pietquin, ``Generalization in mean field games by learning master
  policies,'' in \emph{Proceedings of the AAAI Conference on Artificial
  Intelligence}, vol.~36, no.~9, 2022, pp. 9413--9421.

\bibitem{liu2024scalable}
S.~Liu, X.~Chen, and X.~Di, ``Scalable learning for spatiotemporal mean field
  games using physics-informed neural operator,'' \emph{Mathematics}, vol.~12,
  no.~6, p. 803, 2024.

\bibitem{huang2024unsupervised}
H.~Huang and R.~Lai, ``Unsupervised solution operator learning for mean-field
  games via sampling-invariant parametrizations,'' \emph{arXiv preprint
  arXiv:2401.15482}, 2024.

\bibitem{li2021fourier}
Z.~Li, N.~B. Kovachki, K.~Azizzadenesheli, K.~Bhattacharya, A.~Stuart,
  A.~Anandkumar \emph{et~al.}, ``Fourier neural operator for parametric partial
  differential equations,'' in \emph{International Conference on Learning
  Representations}, 2021.

\bibitem{lu2021learning}
L.~Lu, P.~Jin, G.~Pang, Z.~Zhang, and G.~E. Karniadakis, ``Learning nonlinear
  operators via deeponet based on the universal approximation theorem of
  operators,'' \emph{Nature machine intelligence}, vol.~3, no.~3, pp. 218--229,
  2021.

\bibitem{li2024physics}
Z.~Li, H.~Zheng, N.~Kovachki, D.~Jin, H.~Chen, B.~Liu, K.~Azizzadenesheli, and
  A.~Anandkumar, ``Physics-informed neural operator for learning partial
  differential equations,'' \emph{ACM/JMS Journal of Data Science}, vol.~1,
  no.~3, pp. 1--27, 2024.

\bibitem{liu2025nfmkv}
J.~Liu, L.~Ren, W.~Yao, and X.~Zahng, ``{NF-MKV Net: A} constraint-preserving
  neural network approach to solving {Mean-Field Games} equilibrium,''
  \emph{arXiv preprint arXiv:2501.17450}, 2025.

\bibitem{chen2023hybrid}
X.~Chen, S.~Liu, and X.~Di, ``{A Hybrid Framework of Reinforcement Learning and
  Physics-Informed Deep Learning for Spatiotemporal Mean Field Games},'' in
  \emph{In Proceedings of the 20th International Conference on Autonomous
  Agents and Multiagent Systems}.\hskip 1em plus 0.5em minus 0.4em\relax ACM
  DIgital Library, 2023.

\bibitem{WOS:000378444200012}
T.~A. O'Brien, K.~Kashinath, N.~R. Cavanaugh, W.~D. Collins, and J.~P. O'Brien,
  ``A fast and objective multidimensional kernel density estimation method:
  fastkde,'' \emph{COMPUTATIONAL STATISTICS \& DATA ANALYSIS}, vol. 101, pp.
  148--160, SEP 2016.

\end{thebibliography}

\vspace{12pt}

\end{document}